%% file: main.tex
\documentclass[sigconf,nonacm=true,10pt]{acmart}
\usepackage{graphicx}
\usepackage{caption}
\usepackage{subcaption}
\usepackage{xspace}
\usepackage{enumitem}
\usepackage{multirow}
\usepackage{amsmath}
\usepackage{hyperref}
\usepackage{booktabs}
\usepackage{arydshln}  
\usepackage{array}      
\usepackage{makecell}   
\usepackage{tabularx}
\usepackage{arydshln}  
\renewcommand{\arraystretch}{1.4}
\DeclareUnicodeCharacter{FF0C}{,}
\renewcommand{\arraystretch}{1.5}  

\newcommand{\dmidrule}{%
  \noalign{%
    \vskip\aboverulesep
    \hbox to\linewidth{%
      \leaders\hrule height 0.4pt width 4pt\hfill
    }%
    \vskip\belowrulesep 
  }%
}

\newcommand{\ours}{\textsf{AURA}\xspace}

\begin{document}


\title{AIoT-based Continuous, Contextualized, and Explainable Driving Assessment for Older Adults}




\author{Yimeng Liu$^{1}$, Fangwei Zhang$^{1}$， Maolin Gan$^1$, Jialuo Du$^1$, Jingkai Lin$^1$, Yawen Wang$^1$, Fei Sun$^1$, Honglei Chen$^1$, Linda Hill$^2$, Ruofeng Liu$^1$, Tianxing Li$^1$, Zhichao Cao$^{1}$}
\affiliation{$^1$Michigan State University\country{USA}  \quad $^2$University of California San Diego \country{USA}}








\renewcommand{\shortauthors}{Y. Liu, F. Zhang, M. Gan, J. Du, J. Lin, Y. Wang, F. Sun, H. Chen, L. Hill, R. Liu, T. Li, Z. Cao}

\input{content/abs}



\maketitle

\input{content/1_introduction}

\input{content/2_background}

\input{content/5_observation}

\input{content/3_system_design}

\input{content/6_related_work}

\input{content/7_conclusion}

\clearpage
\bibliographystyle{ACM-Reference-Format}
\bibliography{ref}

\end{document}

%% file: content/abs.tex
\begin{abstract}
The world is undergoing a major demographic shift as older adults become a rapidly growing share of the population, creating new challenges for driving safety. 
In car-dependent regions such as the United States, driving remains essential for independence, access to services, and social participation. 
At the same time, aging can introduce gradual changes in vision, attention, reaction time, and driving control that quietly reduce safety. 
Today’s assessment methods rely largely on infrequent clinic visits or simple screening tools, offering only a brief snapshot and failing to reflect how an older adult actually drives on the road. 
Our work starts from the observation that everyday driving provides a continuous record of functional ability and captures how a driver responds to traffic, navigates complex roads, and manages routine behavior. 
Leveraging this insight, we propose \ours, an Artificial Intelligence of Things (AIoT) framework for continuous, real-world assessment of driving safety among older adults. 
\ours integrates richer in-vehicle sensing, multi-scale behavioral modeling, and context-aware analysis to extract detailed indicators of driving performance from routine trips. 
It organizes fine-grained actions into longer behavioral trajectories and separates age-related performance changes from situational factors such as traffic, road design, or weather. 
By integrating sensing, modeling, and interpretation within a privacy-preserving edge architecture, \ours provides a foundation for proactive, individualized support that helps older adults drive safely. 
This paper outlines the design principles, challenges, and research opportunities needed to build reliable, real-world monitoring systems that promote safer aging behind the wheel.
\end{abstract}

%% file: content/1_introduction.tex
\section{Introduction}
\label{sec:intro}

\begin{figure}[b]
  \centering
  \includegraphics[width=\linewidth]{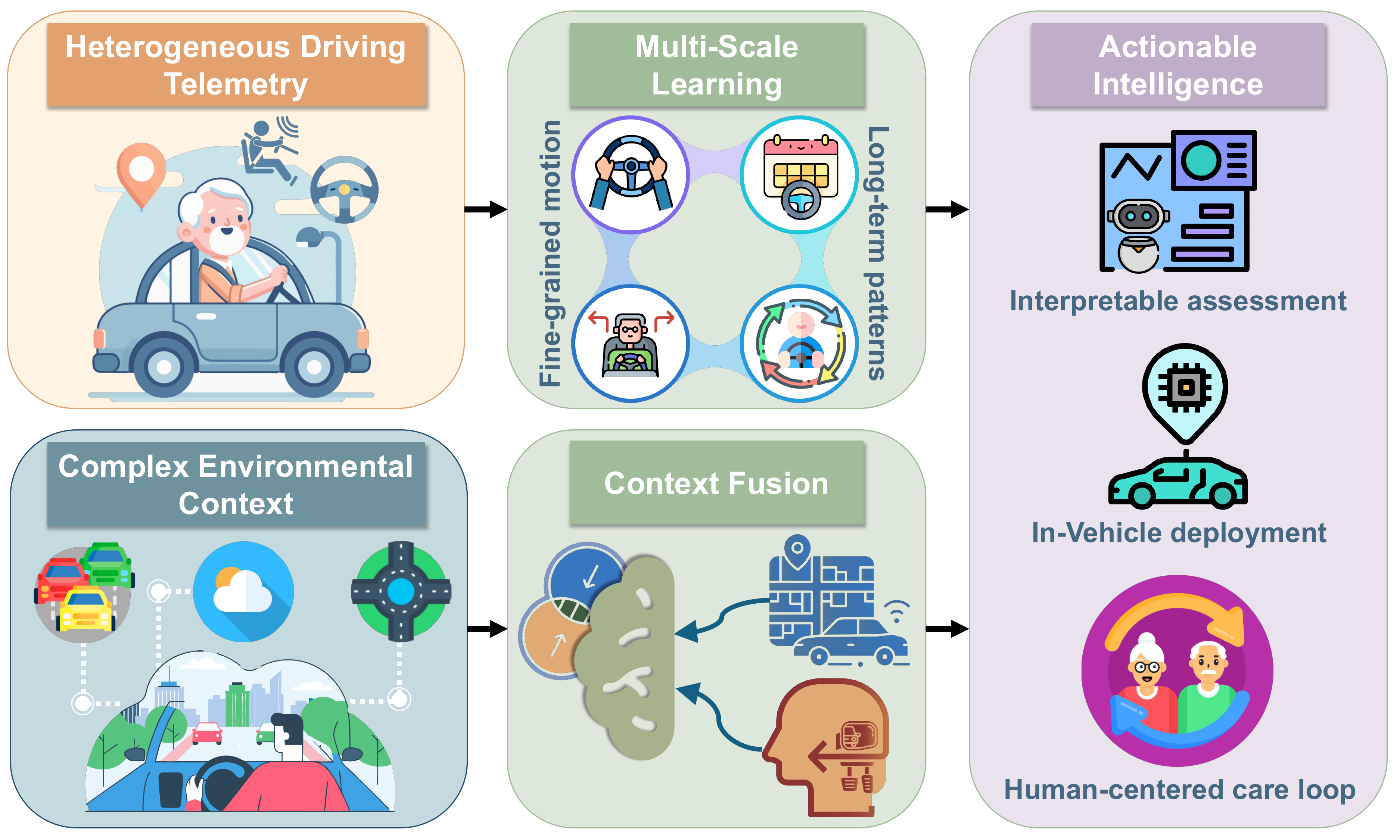}
  \caption{Overview of \ours}
  \label{fig:teaser}
\end{figure}

Aging is one of the most significant global demographic shifts of the 21st century \cite{WHO2024AgeingHealth}. By 2030, one in six people worldwide is projected to be 60 years or older~\cite{WHO2024}. 
This transformation has a direct impact on transportation systems, particularly in car-dependent countries like the United States, where personal driving is deeply embedded in daily life.
In the United States, driving plays a central role in supporting independence, access to essential services, and social engagement~\cite{Molnar2015SelfRegulation, AnsteyDrivingPerformance2022}.
It supports daily needs, personal autonomy, and social participation~\cite{Molnar2015SelfRegulation}. 
It is also closely tied to independence and identity~\cite{Molnar2015SelfRegulation, Carr2006OlderAdultDrivers, YamamotoImpactOfAging2021}. 
Losing the ability to drive can lead to isolation, depression, and faster cognitive or physical decline~\cite{Carr2006OlderAdultDrivers}. 
Thus, the number of older drivers has continued to grow.
Among adults aged 65 and older, licensure rose from 78\% in 2001 to 89\% in 2021~\cite{NHTSAOlderDrivers}; among adults aged 85 and older, it increased from 50\% in 2000 to 59\% in 2020~\cite{FHWA2021DL20}.

At the same time, the aging adult population faces increasing challenges in driving.
In 2022, collisions involving older drivers accounted for over 9,100 fatalities and 270,000 injuries~\cite{CDC2025OlderAdultDrivers}. 
These elevated risks are caused mainly by age-related cognitive decline, which weakens attention, planning, and hazard anticipation~\cite{Davis2020DrivingBehavior, CraneCognitiveFunction2016}. 
It also leads to secondary losses in vision, driving coordination, and situational awareness, further increasing the likelihood of driving errors~\cite{CDC2024CommonInjuries, RapoportCognitionAndDriving2009, BabulalANaturalisticStudy2017}.
Although fully autonomous driving can help prevent seniors from such risks, it remains far from reliable, especially in the complex, socially negotiated environments that older adults encounter. 
As a result, seniors must continue to rely on their own driving abilities, which poses a new challenge: \emph{how to assess seniors' driving ability accurately?}



To solve this problem, state-level policies focus on age-based renewals, medical checks, and post-incident evaluations~\cite{CaliforniaDMV_SeniorDrivers, OregonDMV_50PlusDriverReporting, VA_DrivingFitness_2021, FLHSMV_MatureDriver_RenewalOptions}.
These approaches assume a uniform aging process and provide sparse, reactive assessments that overlook the wide variation in driving behavior and individual health conditions \cite {OttALongitudinalStudy2008}.
To improve the baseline, data-driven approaches across mobile sensing, transportation, and clinical domains have expanded observational capabilities. 
In-vehicle driving monitoring systems detect distraction, physiological states, and occupant conditions~\cite{wang2021distracted, zheng2020v2ifi, umusic2021, MoshfeghiInVehicleSensing2023, TappenInVehicleSensors2023}, and naturalistic driving studies link behavior to cognitive functional change~\cite{Di2023, DiUsingNaturalistic2021, Vardaki2016, AraGhoreishi2023, Bayat2021}.
%

 To integrate insights from in-vehicle driving-monitoring systems and naturalistic driving studies, we design a driving simulation testbed using the CARLA~\cite{carla}. 
It can capture precise body movement, detailed driving actions, and complete environmental context in a synchronized, high-resolution manner. 
This dataset expanded observations of how older adults actually drive, moment-to-moment. 
When we analyze these signals and large real-world datasets, including AAA LongROAD and DRIVES (detailed in \S \ref{sec:obs}),  we reveal several limitations and fundamental challenges as follows:
\begin{itemize}
    \item Current datasets cannot cleanly capture senior driving patterns because real-world trips occur under highly varied environments. 
    Also, driving signals combined with strong individual self-regulation produce highly personalized driving signatures that obscure stable features.

    \item Driving behavior is tightly shaped by environmental conditions. 
    However, current datasets lack reliable information about traffic, road geometry, and weather conditions. 
    Without this contextual data, many behavior changes become ambiguous, and models cannot distinguish environment-driven variability from meaningful functional decline.

    \item Existing black-box models may flag anomalies but offer little explanatory value, preventing clear connections between observed behavior and underlying cognitive processes. 
    Even with more abundant data, current methods still cannot provide a clear interpretation of safe driving.
\end{itemize}
Together, these issues reveal a fundamental gap: \emph{current data-driven assessment designs can collect large volumes of driving data but still cannot reliably identify early changes in driving capability in real-world settings, posing significant risks to older drivers.}

This paper presents a scientific and system vision for supporting seniors' safe driving and outlines \ours, for \textbf{A}ging \textbf{U}nderstanding \& \textbf{R}isk \textbf{A}ssessment, shown in Figure \ref{fig:teaser}.
It is an Artificial Intelligence of Things (AIoT) based driving assessment system. 
\ours provides continuous insight into the evolution of driving behavior, perceptual responses, and human motion. 
By incorporating richer and privacy-preserving sensing systems, it captures both vehicle control patterns and subtle in-cabin behavioral features. 
Also, the multimodal, context-aware integration enables the system to interpret driving actions in relation to environmental conditions, producing a more accurate, situationally grounded assessment. 
Finally, \ours emphasizes Explainable and Actionable AI.
It requires the system to operate entirely within the vehicle to preserve privacy and also translate complex behavioral signals into clear guidance that drivers, families, and clinicians can use within a human-centered care loop.

\vspace{1mm}
\noindent
\textbf{Our Contributions.}
In this vision paper, we outline a research roadmap for developing \ours:

\begin{itemize}
\item We design a testbed that records multimodal, high-fidelity signals to reflect how older adults actually drive and expose the gaps in previous methods.
\item We propose a continuous, privacy-preserving framework for monitoring naturalistic driving behavior through non-intrusive in-vehicle sensing.
\item We distill three grand challenges: reliable capability-driven behavior modeling, context-aware understanding, and explainable decision-making, that define the technical and ethical core of senior driving assessment.
\item We propose three future research directions that go beyond \ours: Elderly Cognitive World Model, Federated Embodied Learning, and Digital Therapeutics and Active Health.
\end{itemize}

\vspace{1mm}
\noindent \textbf{Opportunities for the AIoT Community.}
\ours reveals a research gap and opens new opportunities for the AIoT research community in sensing, processing, and AI. 
For mobile and wireless sensing, our work demonstrates the need for richer, privacy-preserving modalities capable of detecting micro-movements and fine-grained driving behavior. 
For the AI community, our findings reveal that senior driving is an ideal setting for developing causal, interpretable, and longitudinal modeling pipelines that move beyond black-box classifiers and explain why behavior changes over weeks or months. 
Together, these requirements position senior driving safety as a cross-disciplinary AIoT challenge in which embedded sensing, edge computing, and trustworthy machine learning must collaborate to deliver reliable, human-centered support.

%% file: content/2_background.tex
\section{Background}
From the broader motivation, this section now examines why aging affects driving safety. 
We begin by outlining the key difficulties seniors face on the road, emphasizing cognitive decline as a primary factor in perception, decision-making, and control. 
We then survey existing regulatory, clinical, and data-driven assessment frameworks to highlight their limitations in capturing real-world driving ability.
This context reveals a widening gap between the complexities of age-related decline and the episodic, low-resolution evaluation methods used today, underscoring the need for continuous, context-aware monitoring.

\subsection{Cognitive Decline as the Core Issue}
\label{sec:cogn_decline}
Seniors face increasing difficulty on the road primarily because of age-related cognitive decline \cite{anstey2011chronological, RapoportCognitionAndDriving2009}.
These declines translate into measurable changes on the road. 
Neurodegenerative conditions such as Mild Cognitive Impairment (MCI) and Alzheimer’s disease disrupt these functions, leading to reduced situational awareness, slower decisions, and less consistent control~\cite{fuermaier2019driving, Wagner2014CognitionDriving}.
Increased speed variability often reflects reduced attention, hesitation during turns or merges signals weakened executive function, and lane-keeping drift indicates impaired visuospatial processing~\cite{Wagner2014CognitionDriving, KrasniukDrivingErrorsPredictive2022}.
Because these changes emerge gradually and are often masked by driving experience or self-regulation, they are challenging to detect using infrequent clinical visits or periodic licensing checks.



\subsection{Driving Assessment}
Efforts to assess and regulate older adults’ driving fitness currently span three domains: policy, clinical practice, and data-driven analytics.
Each addresses distinct aspects of the problem, but collectively they fail to provide a scalable, continuous, and personalized solution. 
However, these approaches lack integration, longitudinal depth, and contextual sensitivity, leading to fragmented assessments.

\noindent \textbf{Policy-Based Regulation.}  
Governmental policies across U.S. states provide the primary regulatory framework for managing driving safety among older adults.
However, these frameworks rely primarily on episodic, age-triggered evaluations rather than individualized assessment. 
States differ substantially in their approaches: 
California conducts vision testing at renewals after age 70 ~\cite{CaliforniaDMV_SeniorDrivers}; 
Oregon requires physicians to report medical conditions that may impair driving, but otherwise follows uniform renewal procedures from age 50~\cite{OregonDMV_50PlusDriverReporting}; 
Virginia mandates in-person renewal and vision screening for drivers 75 and older~\cite{VA_DrivingFitness_2021}; 
and Florida requires only a vision test for drivers over 80~\cite{FLHSMV_MatureDriver_RenewalOptions}. 
Although these policies aim to maintain road safety, they rely on broad proxies such as age, medical reporting, or isolated screening events.

\noindent \textbf{Clinical Assessment.}  
Clinical evaluations, primarily cognitive screening tests and structured on-road examinations, constitute the conventional approach to determining driving fitness in older adults. 
Screening tools such as the Mini-Mental State Examination (MMSE) and Montreal Cognitive Assessment (MoCA) are used to detect a general cognitive profile.
However, they fail to detect driving-specific cognitive demands.
Additionally, they have limited ability to predict performance in the dynamic, multitasking environment of real-world driving~\cite{BabulalANaturalisticStudy2017, FuermaierDrivingFitness2018, OttTheNeuroscience2010}, making it challenging to assess senior driving ability accurately. 
On-road tests provide direct observation but are resource-intensive, difficult to standardize, and capture only a brief snapshot of behavior~\cite{Akinwuntan2019OnRoadDrivingTests}. 
Both methods occur at low frequency and outside natural driving contexts, making them insensitive to gradual cognitive changes or day-to-day variability~\cite{Davis2012RoadTestNaturalistic}. 

\noindent \textbf{Data-Driven Approaches.}  
Recent research has expanded the ways in which driving performance can be assessed across sensing, simulation, and naturalistic data modalities. 
In-car sensing systems focus on monitoring driver state rather than driving competence: smartphone-based sensors detect distraction~\cite{wang2021distracted}, RF and UWB systems track physiological status and fine-grained occupant motion~\cite{zheng2020v2ifi, umusic2021}, and acoustic sensing supports coarse, continuous monitoring. 
These methods provide rich moment-to-moment observations but do not directly evaluate driving ability. 
Naturalistic driving datasets capture long-term, real-world behavior and reveal associations between everyday patterns and functional decline~\cite{Di2023, Vardaki2016, AraGhoreishi2023, Bayat2021}. 
However, these studies primarily use black-box models.
Without transparent reasoning that explains why specific behaviors signal elevated risk, clinicians and regulators cannot justify interventions such as screening, restricted licensing, or driving cessation \cite{Yassuda01011997, anstey2011chronological}.
Building on these observational tools, emerging AI-based systems aim to infer cognitive health directly from driving traces~\cite{DiUsingNaturalistic2021, ZhangDetectingMildCognitive2023, HasanTheRoleOf2024, GomezTheRoleOf2024, AnonymousAIDrivenEarly2024, SendakFromDevelopment2024}.

\noindent \textbf{Motivation.}
Although government programs, clinical evaluations, and research prototypes recognize the importance of senior driving safety, these approaches rely on infrequent screenings and controlled tests, which miss how cognitive demands unfold in daily driving. 
The earliest signs of decline appear not in clinics but in routine behavior, subtle changes during familiar commutes, moment-to-moment decisions, and natural interactions with traffic and the environment. 
This gap motivates a new direction: continuous monitoring grounded in real-world driving behavior, enriched with contextual awareness, and supported by interpretable reasoning that families, clinicians, and regulators can trust. 
Our objective is to move beyond episodic assessments toward systems that reflect the true cognitive challenges seniors face on the road, delivering earlier, actionable, and transparent insights into driving fitness.

%% file: content/5_observation.tex
\begin{table*}[ht]
\centering
\caption{\textbf{Summary of Naturalistic and Controlled Driving Datasets}}
\label{tab:driving_datasets_full_page_en}

\renewcommand{\arraystretch}{1.25}
\setlength{\aboverulesep}{0.6ex}
\setlength{\belowrulesep}{0.6ex}

\newcolumntype{C}[1]{>{\centering\arraybackslash}m{#1}}

\begin{tabular}{C{2.6cm} C{3cm} C{2.4cm} C{2.2cm} C{5.8cm}}
\specialrule{2pt}{0pt}{0pt}
\textbf{Dataset} &
\textbf{Sample Size} &
\textbf{Age Range} &
\textbf{Time Span} &
\textbf{Key Features} \\
\specialrule{2pt}{0pt}{0pt}
\vspace{1.5em}
\textbf{LongROAD} &
\vspace{1.5em}
2402 &
\vspace{1.5em}
65--80+ &
\vspace{1.5em}
5 years &
\makecell[l]{
• Population Diversity \\
• Trip-level driving logs \\
• Time, mileage, and speed records 
}
\\
\dmidrule
\vspace{0.5em}
\textbf{DRIVES} &
\vspace{0.5em}
414 &
\vspace{0.5em}
65--90+ &
\vspace{0.5em}
3 months &
\makecell[l]{
• High-resolution GPS traces \\
• Cognitive assessment data
}
\\
\dmidrule

\textbf{CARLA Simulator} &
\makecell[c]{20 
\\ 
(10 seniors, 10 youth)
} &
18--80+ &
\vspace{-0.5em}
One session &
\vspace{-2em}
\makecell[l]{
• High-frequency telemetry data \\
• Full body tracking \\
• Controlled driving scenarios
}
\\

\specialrule{2pt}{0pt}{0pt}
\end{tabular}
\end{table*}

\section{Observations and Insights}
\label{sec:obs}

This section presents empirical insights from three complementary datasets: a controlled CARLA-based testbed, the high-resolution DRIVES dataset, and the population-scale LongROAD real-world dataset. 
Together, these enable us to examine senior driving behavior across progressively richer, more realistic settings. 
We first describe the datasets and experimental scope, then analyze how raw driving signals reveal distinct senior behavioral patterns, how these behaviors change with environmental context, and how cognitive and functional factors shape the observed differences shown in Table \ref{tab:driving_datasets_full_page_en}. 
Finally, we highlight why these patterns are difficult to interpret reliably, motivating the challenge that follows.

\subsection{Datasets and Experimental Scope}
\begin{figure}[b]
  \centering
  \includegraphics[width=\linewidth]{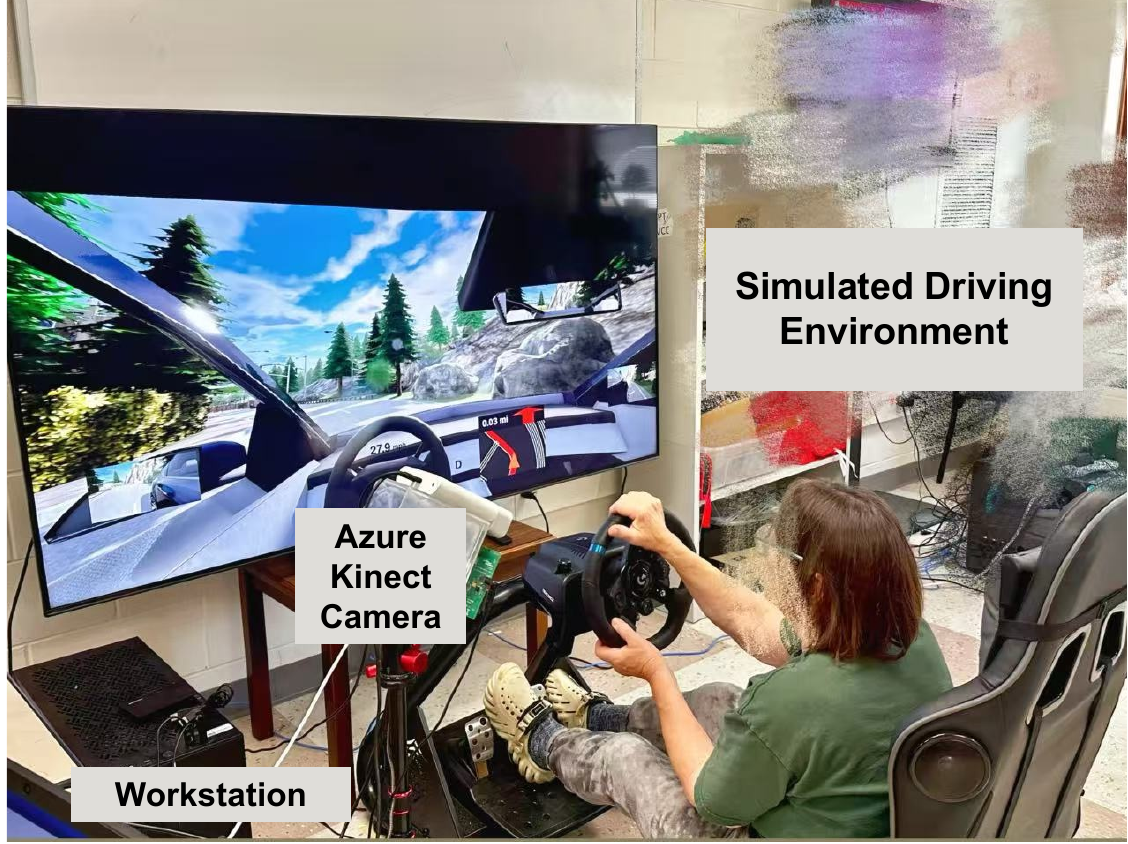}
  \caption{Overview of the CARLA simulator environment employed for controlled driving data collection.}
  \label{fig:testbed_overview}
\end{figure}

\noindent \textbf{CARLA Simulator Testbed.}
We design a CARLA-based controlled driving testbed that closely approximates real-world driving while providing complete experimental control. 
The setup includes a cockpit-style simulator equipped with a driver's seat, steering wheel, and pedals, designed to replicate the physical feel of a real vehicle, shown in Figure \ref{fig:testbed_overview}.
Beyond that, a wide monitoring display provides an immersive visual field comparable to an in-car windshield view, including all the essential vehicle interfaces, such as rear-view mirrors, speed indicators, and navigation displays.
It also supports realistic road environments, including multilane urban streets, highways, intersections, traffic lights, and stop signs, with configurable weather and lighting to match real-world variability.
The platform records all relevant signals at high resolution, including steering angle, throttle and brake inputs, speed profiles, vehicle state, and surrounding traffic. 
In addition, an Azure Kinect camera captures full-body motion, enabling detailed behavioral analysis.
Together, this controlled and realistic environment enables repeatable experiments, precise measurement, and systematic evaluation of the driving behavior.

We conduct an IRB-approved experiment with 10 seniors aged 65 and above and 10 younger adults to participate in our controlled driving experiment.
To reduce bias arising from unfamiliarity with simulated environments, each participant first completed a guided tutorial drive on an easy route to become familiar with the simulator's vehicle dynamics and visual interface.
We design a fixed driving environment and predefined route that includes scripted traffic, pedestrian activity, and two controlled weather phases: sunny conditions at the beginning and rain near the end. 
All environmental parameters are recorded and replayed deterministically, ensuring that every participant experiences the same scenario across runs. 
Each participant must follow on-screen navigation instructions to complete the entire route, thereby simulating a real-world driving scenario. 


\begin{figure*}[!ht]
   \centering
   \begin{subfigure}[b]{0.235\textwidth}
       \centering\includegraphics[width=\textwidth]{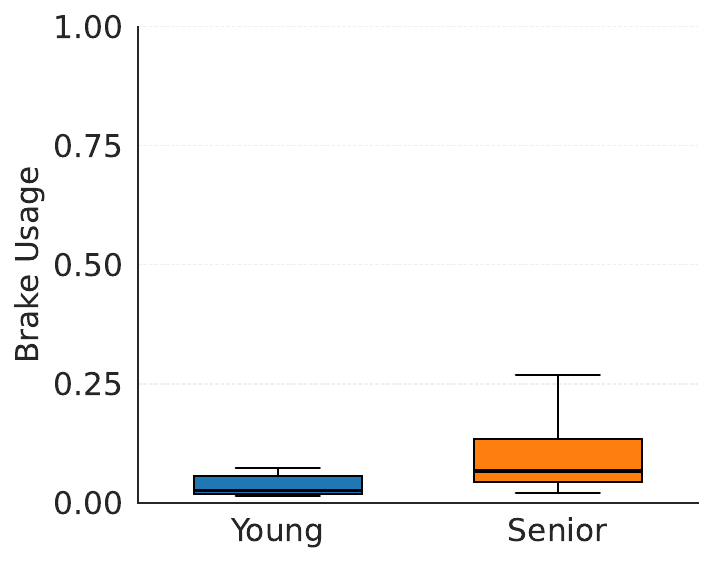}
       \caption{Brake Usage}
   \end{subfigure}
   \hfill
   \begin{subfigure}[b]{0.235\textwidth}
       \centering\includegraphics[width=\textwidth]{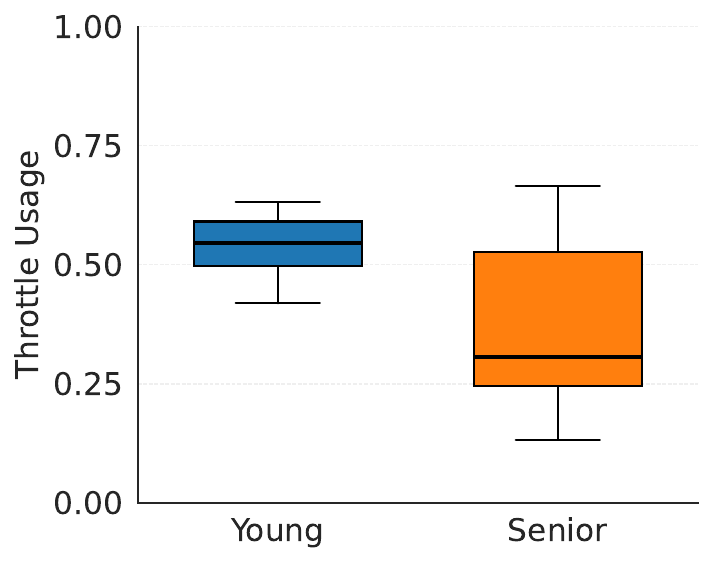} 
       \caption{Throttle Usage}
   \end{subfigure}
   \hfill
   \begin{subfigure}[b]{0.235\textwidth}
       \centering\includegraphics[width=\textwidth]{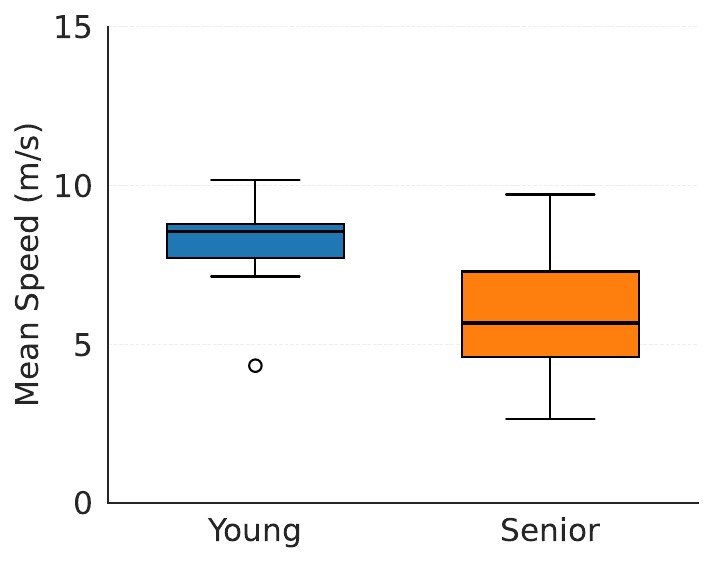}
       \caption{Speed}
   \end{subfigure}
   \hfill
   \begin{subfigure}[b]{0.235\textwidth}
       \centering\includegraphics[width=\textwidth]{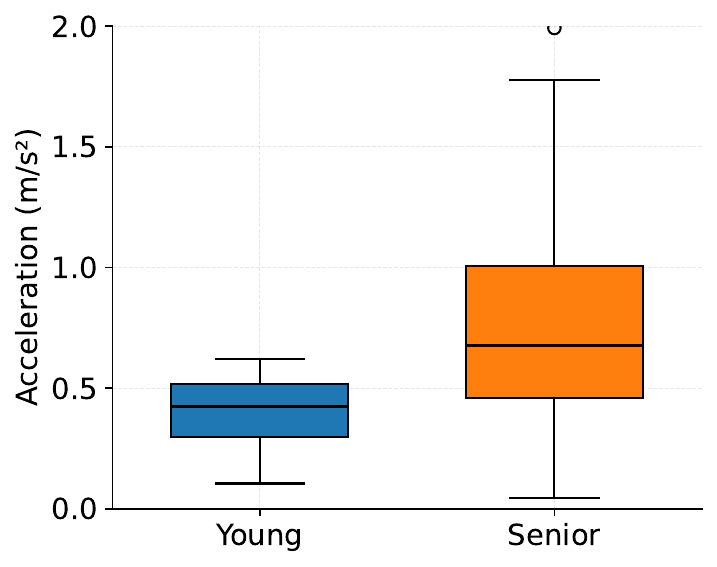} 
       \caption{Acceleration}
   \end{subfigure}
    \caption{Driving behavior is different between young and senior drivers.}
        \label{fig:tb-ys}
\end{figure*}

\noindent \textbf{Longitudinal Research on Aging Drivers (LongROAD) \cite{LongROAD}.}
The LongROAD dataset is a large-scale, multi-site longitudinal cohort comprising 2,402 older adults, tracked over 5 years, with 12,361 assessment records across repeated study visits. 
Its primary strength lies in its demographic and socioeconomic diversity, including 1,277 women and 1,125 men, spanning three major age groups from 65 to over 80, and reflecting broad racial and ethnic representation, including White, Black, Asian, American Indian, Hispanic/Latino, and other minority groups. 
Education levels range from high school or less to advanced graduate degrees, and marital status includes married, widowed, divorced, and never married.
Another key feature of LongROAD is its integration of rich multi-year driving data. 
Each participant will record trip counts, daily and weekly driving distance, time-of-day distributions, speed profiles, hard-braking events, and aggressive turning signatures for 5 years. 

\medskip
\noindent \textbf{DRIVES \cite{DRIVE}.}
The Washington University DRIVES dataset provides dense, high-resolution naturalistic driving telemetry for 414 older adults, covering 25,655 trips and more than 878,381 miles of real-world driving. 
Participants completed an average of 61.97 trips, with individual totals ranging from 1 to 92 visits and mileage spanning from a few miles to 1255 miles.
Each trip includes detailed summaries, with a mean duration of 74.8 minutes, and second-by-second OBD-II and GPS breadcrumbs that capture speed, acceleration, braking, and full geospatial traces.
These rich signals enable fine-grained reconstruction of driving behaviors, including speed regulation, braking aggressiveness, turning forces, mobility routines, and temporal driving rhythms.
A key feature of DRIVES is its unique integration of comprehensive neuropsychological assessments, including the Trail Making Tests, verbal fluency, memory recall, and the Clinical Dementia Rating (CDR). 
This coupling enables direct analysis of how cognitive changes unfold alongside subtle longitudinal shifts in driving behavior.

\subsection{Senior Driving Pattern}
\label{sec:challenge1}

\textbf{What distinctive behaviors define senior driving?}\\
These questions start our investigation into the complex interplay between aging and everyday driving behavior. 
Driving is a vibrant but multifaceted activity that reflects motor control, decision-making, accumulated experience, and adaptive self-regulation. 
Our goal in this work is to disentangle these factors by systematically characterizing patterns among senior drivers and identifying behavioral features that meaningfully signal increased safety risk in naturalistic settings.

\begin{figure}[t]
   \centering
   \begin{subfigure}[b]{0.235\textwidth}
       \centering\includegraphics[width=\textwidth]{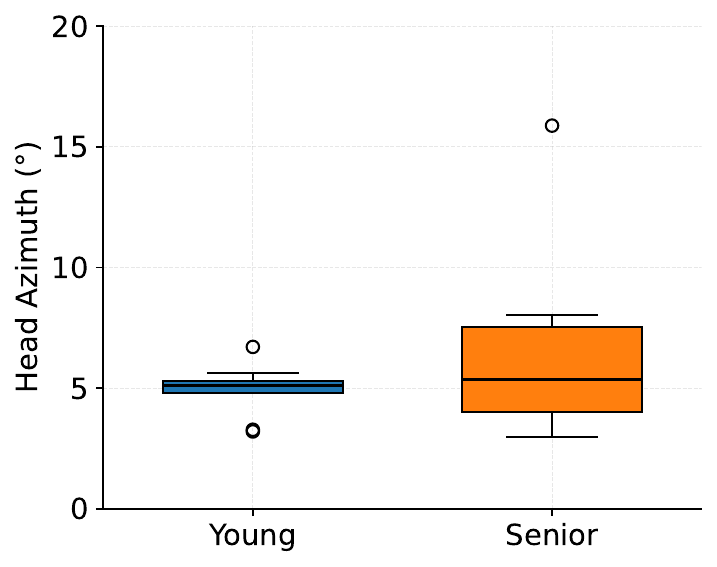}
       \caption{Azimuth Angle Change}
   \end{subfigure}
   \hfill
   \begin{subfigure}[b]{0.235\textwidth}
       \centering\includegraphics[width=\textwidth]{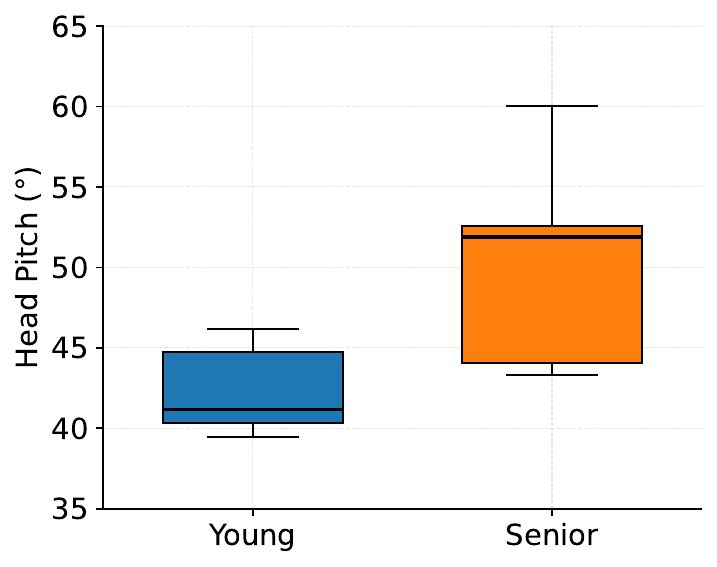} 
       \caption{Elevation Angle Change}
   \end{subfigure}
    \caption{Head movement with age difference. }
    \label{fig:t-body}
\end{figure}

\begin{figure*}[ht]
   \centering
   \begin{subfigure}[b]{0.23\textwidth}
       \centering\includegraphics[width=\textwidth]{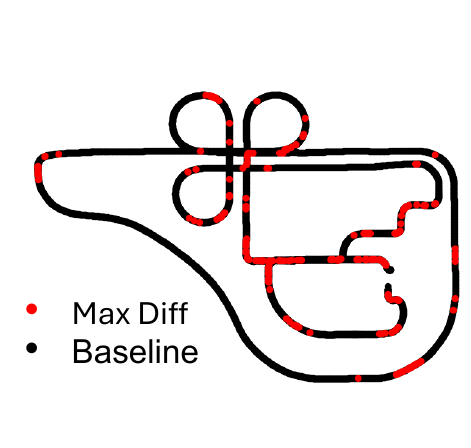}
       \caption{Composite}
   \end{subfigure}
   \hfill
   \begin{subfigure}[b]{0.23\textwidth}
       \centering\includegraphics[width=\textwidth]{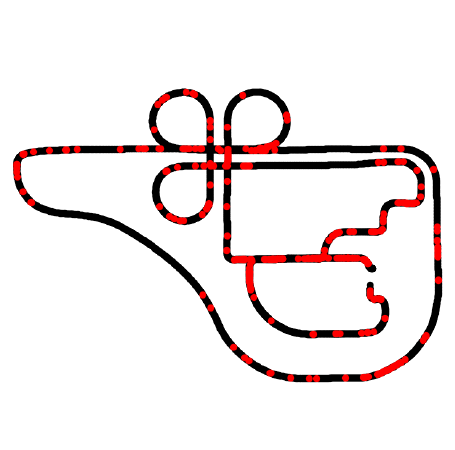} 
       \caption{Stability}
   \end{subfigure}
   \hfill
   \begin{subfigure}[b]{0.23\textwidth}
       \centering\includegraphics[width=\textwidth]{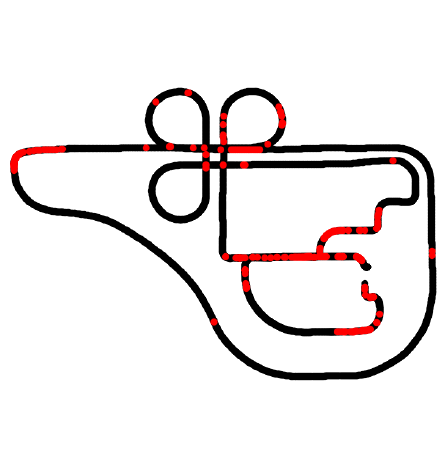} 
       \caption{Reaction}
   \end{subfigure}
   \hfill
   \begin{subfigure}[b]{0.23\textwidth}
       \centering\includegraphics[width=\textwidth]{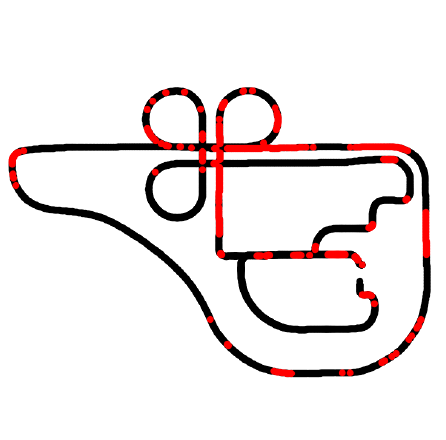} 
       \caption{Route Deviation}
   \end{subfigure}
    \caption{Route-level safety score differences between senior and young drivers. Red points mark locations with maximal group divergence, while black points indicate baseline similarity. Senior drivers show widespread instability and reduced fluidity, with delayed reactions at intersections and larger route deviations in curves.}
        \label{fig:t_map}
\end{figure*}

\noindent \textbf{CARLA Simulator Observation.}
We isolated age-related driving patterns in the simulated environment.
We record data at 1 Hz and analyze how general driving behaviors differ systematically among senior drivers.
As shown in Figure~\ref{fig:tb-ys}, seniors consistently demonstrate a slower increase in brake usage and reduced throttle.
They will control the mean and acceleration more slowly than youth.
These results suggest a risk-averse control profile, consistent with compensatory adjustments for slower processing or diminished confidence in dynamic interactions. 

Beyond driving performance, our body-tracking data provides additional insight into how drivers gather visual information while navigating. 
We compute frame-to-frame changes in head angle to quantify the frequency and magnitude of head movements. 
As shown in Figure~\ref{fig:t-body}, older adults exhibit more frequent and wider-range head-scanning motions compared to younger drivers. 
This pattern suggests an adaptive strategy: seniors compensate for age-related reductions in visual acuity and divided attention by increasing head and eye movements to maintain adequate situational awareness throughout the drive.



We further investigate senior driving safety in the context of route-level safety score analysis. 
This score integrates four dimensions: driving stability, reaction speed, and route deviation. 
These dimensions respectively reflect how steadily a driver maintains control, how quickly decisions about the next turn are made, and how consistently a driver follows the intended route. 
The four sub-scores are computed as follows:
\begin{align}
S_{\text{stab}}(t) &= 1 - \frac{\sigma_{\text{steer}}(t)}{\max_u \sigma_{\text{steer}}(u)}, \\
S_{\text{react}}(i) &= \alpha, S_{\text{time}}(i) + (1-\alpha), S_{\text{fluent}}(i), \\
S_{\text{route}}(i) &= 1 - \frac{D_i}{\max_k D_k}.
\end{align}

Here, $t$ denotes each one-second telemetry sample and $i$ denotes maneuver events (e.g., turns or braking). 
$\sigma_{\text{steer}}(t)$ is the local variance of steering angle, indicating lane-keeping stability. 
$F_{\text{raw}}(t)$ captures abrupt control fluctuations across throttle, brake, and steering, reflecting operational smoothness. 
$S_{\text{time}}(i)$ and $S_{\text{fluent}}(i)$ are normalized reaction-time and maneuver-smoothness scores, combined by weight $\alpha$ to characterize event-level responsiveness. 
$D_i$ measures deviation from the planned route during event $i$. 
Together, these parameters summarize stability, fluidity, responsiveness, and navigation accuracy in a compact behavioral representation.

Based on the above scoring metrics, we compute each score for each route point.
We then aggregate the senior and driver cores, identify the route segments, and determine the route segments. 
As shown in Figure~\ref{fig:t_map}, red markers denote points with maximal divergence. 
The results reveal that seniors differ substantially from younger drivers across multiple dimensions. 
The stability gaps appear throughout nearly the entire route, indicating more frequent micro-corrections and less smooth control. 
Reaction-time differences are particularly pronounced at intersections, where seniors systematically delay their responses when initiating turns. 
Route deviation differences are concentrated around curves, indicating that older adults tend to drift farther from the intended path. 
Together, these findings highlight route-specific stress points at which age-related behavioral changes are most salient.

\noindent \textbf{Remarks.}
Our findings reveal consistent, measurable differences in the driving behavior of older adults. 
They use lower throttle, brake more abruptly, and maintain lower speed and acceleration.
Also, they exhibit broader and more frequent head-scanning motions.
These distinctive patterns constitute a clear senior driving signature, indicating that aging reshapes both the vehicle's moment-to-moment control and the broader habits that develop over time.

\begin{figure*}[t]
   \centering
   \begin{subfigure}[b]{0.235\textwidth}
       \centering\includegraphics[width=\textwidth]{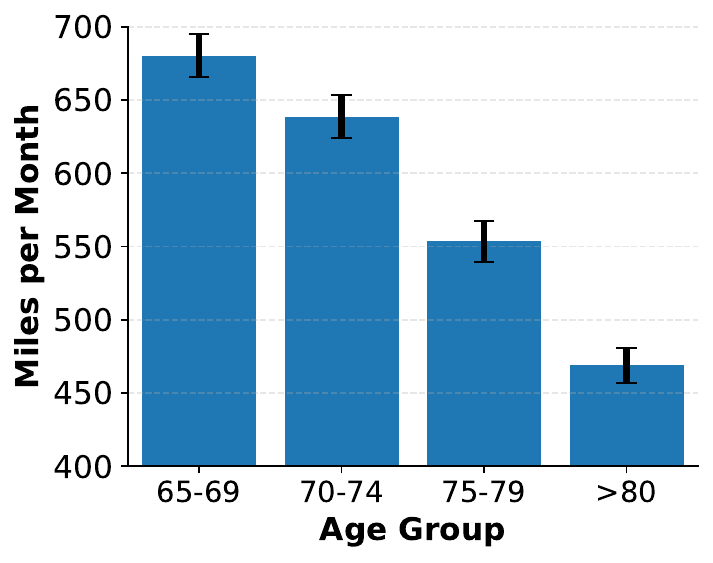}
       \caption{Miles Per Month}
    \label{fig:l_m}
   \end{subfigure}
   \hfill
   \begin{subfigure}[b]{0.235\textwidth}
       \centering\includegraphics[width=\textwidth]{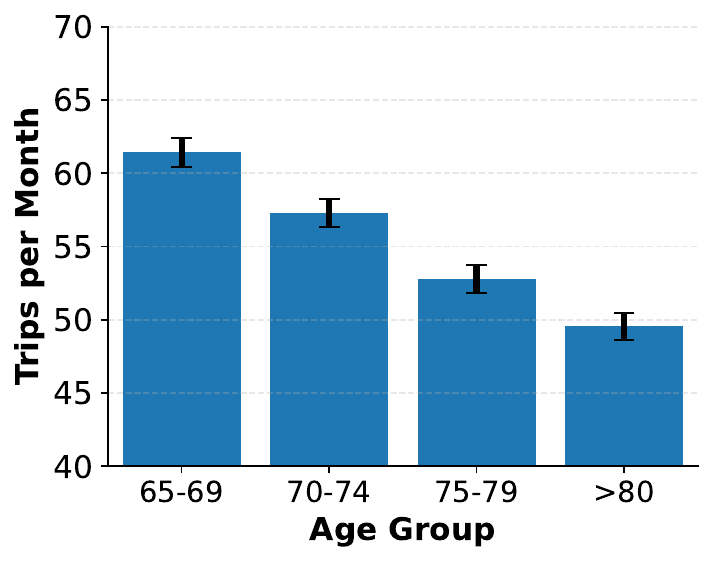} 
       \caption{Trips per Month}
        \label{fig:l_t}
   \end{subfigure}
   \hfill
   \begin{subfigure}[b]{0.235\textwidth}
       \centering\includegraphics[width=\textwidth]{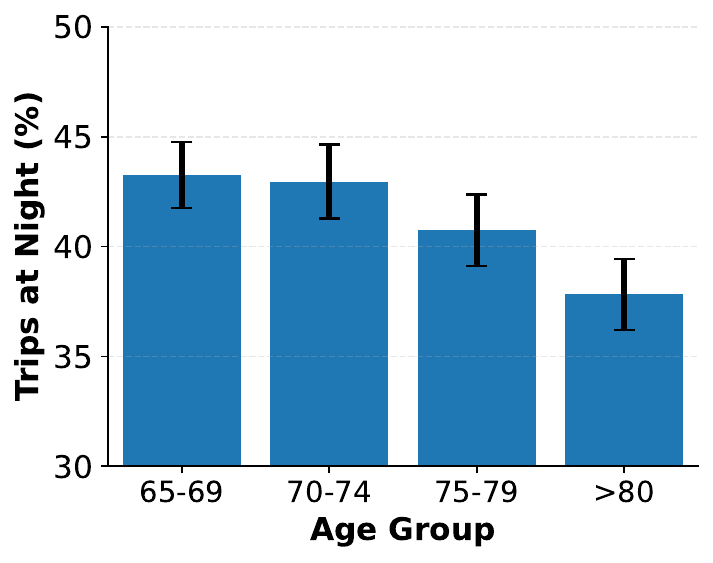}
       \caption{Trips at Night}
    \label{fig:l_night}
   \end{subfigure}
   \hfill
   \begin{subfigure}[b]{0.235\textwidth}
       \centering\includegraphics[width=\textwidth]{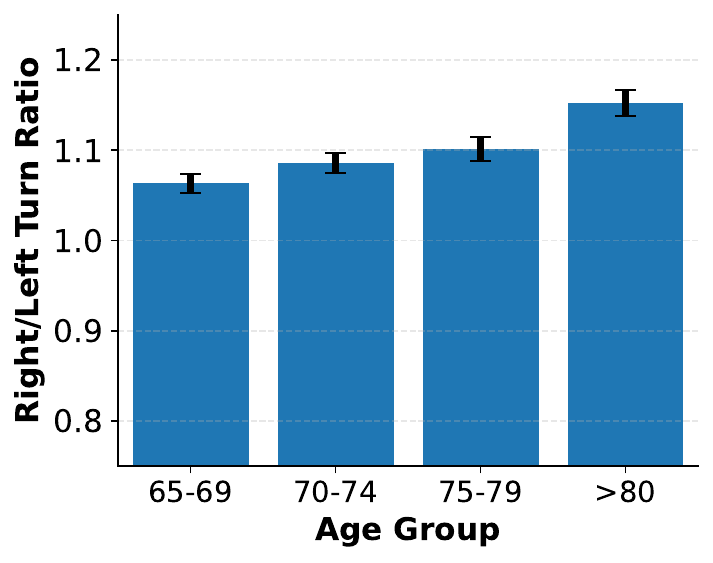} 
       \caption{Right/Left Turn Ratio}
        \label{fig:l_rl}
   \end{subfigure}
    \caption{LongROAD dataset: different age-group driving patterns. }
        \label{fig:l_age}
\end{figure*} 

\medskip
\noindent  \textbf{What is senior real-world driving behavior?}
Laboratory studies conducted in controlled environments help reveal specific age-related driving patterns.
However, their real-world driving features are not clearly demonstrated.
We aim to move beyond the laboratory and comprehensively examine senior driving patterns.

\noindent \textbf{LongROAD Observation.} 
We analyzed the LongROAD dataset, which records years of real-world driving behavior from thousands of older adults. 
Figure~\ref{fig:l_age} summarizes key age-dependent mobility patterns. 
Both total monthly driving distance and trip frequency decline steadily with age, with miles traveled per month dropping by more than 50\% between drivers aged 65–74 and those over 80. 
The fraction of trips taken also shows a marked reduction from ages 61 to 80. 
In addition, older adults are increasingly reluctant to drive at night: the proportion of night trips declines markedly, with a particularly pronounced drop among drivers aged 65–74 and those aged 80 and older. These trends reflect strong self-regulation behaviors, as seniors progressively limit exposure to challenging conditions to preserve safety and comfort.
Additionally, route-level analysis reveals a marked preference for right turns over left turns, reflecting avoidance of high-risk choices that demand rapid judgment under uncertainty. 
These observations provide compelling evidence that older drivers engage in continuous self-regulation—actively shaping when, where, and how they drive to maintain perceived safety. 
Such voluntary adaptation complicates the inference of underlying cognitive decline.

\begin{figure}[t]
   \centering
   \begin{subfigure}[b]{0.235\textwidth}
       \centering\includegraphics[width=\textwidth]{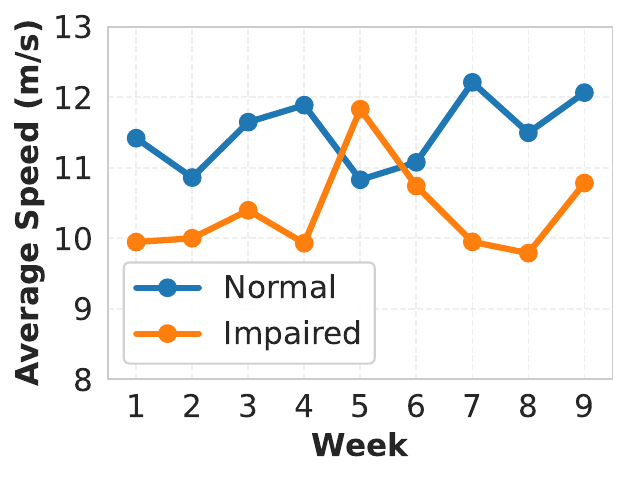}
       \caption{Weekly Speed}
   \end{subfigure}
   \hfill
   \begin{subfigure}[b]{0.235\textwidth}
       \centering\includegraphics[width=\textwidth]{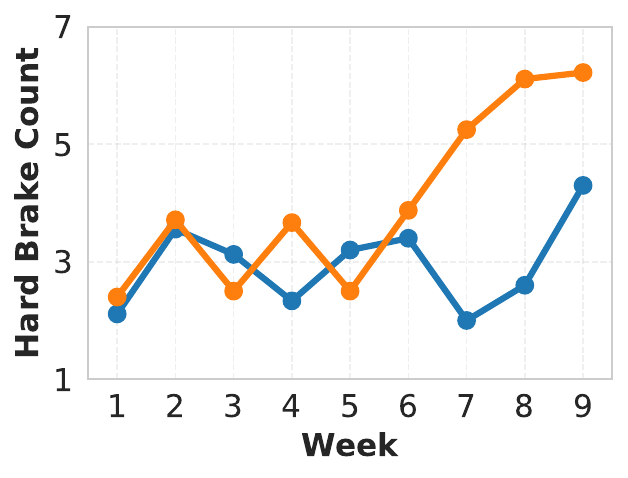} 
       \caption{Weekly Hard Brake}
   \end{subfigure}
    \caption{Senior weekly driving features.}
    \label{fig:d-weekly}
\end{figure}

\noindent \textbf{DRIVE Observation.}
To better understand how aging affects driving beyond general behavioral differences, it is essential to examine the subtle manifestations of cognitive change. 
Cognitive decline, as stated in Section \ref{sec:cogn_decline}, can alter attention, anticipation, and decision-making. 
To explore these nuances, we draw on the DRIVES dataset, which links continuous in-vehicle telemetry with neuropsychological assessments. 
As shown in Figure~\ref{fig:d-weekly}, cognitively normal and impaired seniors exhibit distinct week-to-week patterns in mean speed and hard-braking frequency. 
Impaired drivers brake more often and drive at consistently lower speed patterns suggestive of reactive control rather than anticipatory planning. 
However, we find out that these deviations are not stable across weeks.
Individual trajectories fluctuate substantially, reflecting the interplay between cognitive status, environmental demands, and compensatory driving strategies. 

\begin{figure}[t]
  \centering
  \includegraphics[width=0.55\linewidth]{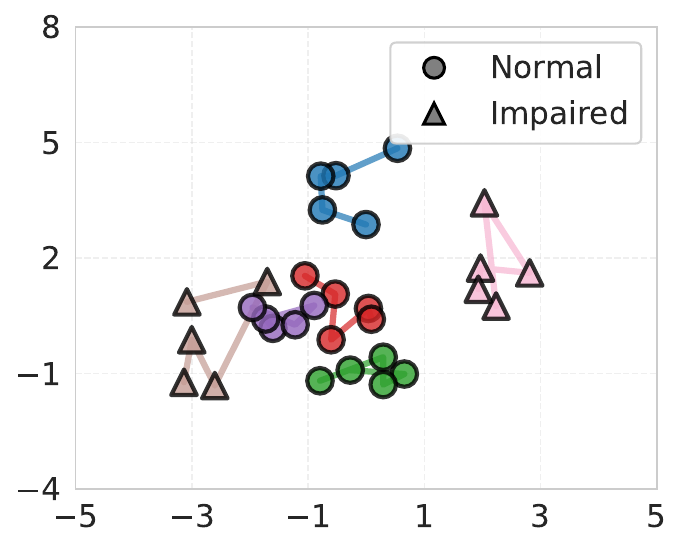}
  \caption{PCA result for the driving behaviors.}
  \label{fig:d-pca}
\end{figure}

We further aggregate participants ' data, including speed, acceleration, trip duration, and turning, and apply Principal Component Analysis (PCA) to examine population-level structure.
As illustrated in Figure~\ref{fig:d-pca}, each color represents a unique driver, with circles denoting cognitively normal seniors and triangles indicating those with mild impairment. 
Notably, the embeddings show minimal separation by cognitive status.
In contrast, the points cluster strongly within individuals, reflecting highly stable, person-specific driving signatures. 
This result highlights a central difficulty in senior-driver analysis: long-practiced driving habits dominate the feature space, making it significantly harder to extract subtle behavioral deviations associated with aging or cognitive change.

\begin{figure*}[t]
   \centering
   \begin{subfigure}[b]{0.235\textwidth}
       \centering\includegraphics[width=\textwidth]{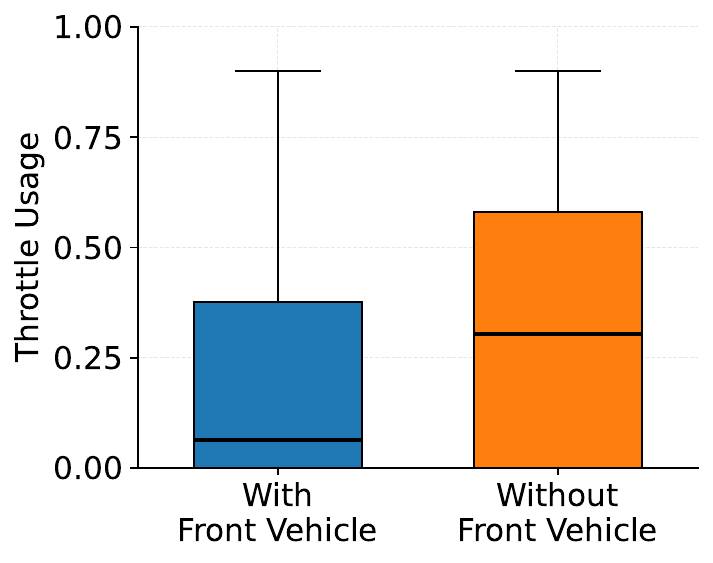}
       \caption{Throttle Usage}
   \end{subfigure}
   \hfill
   \begin{subfigure}[b]{0.235\textwidth}
       \centering\includegraphics[width=\textwidth]{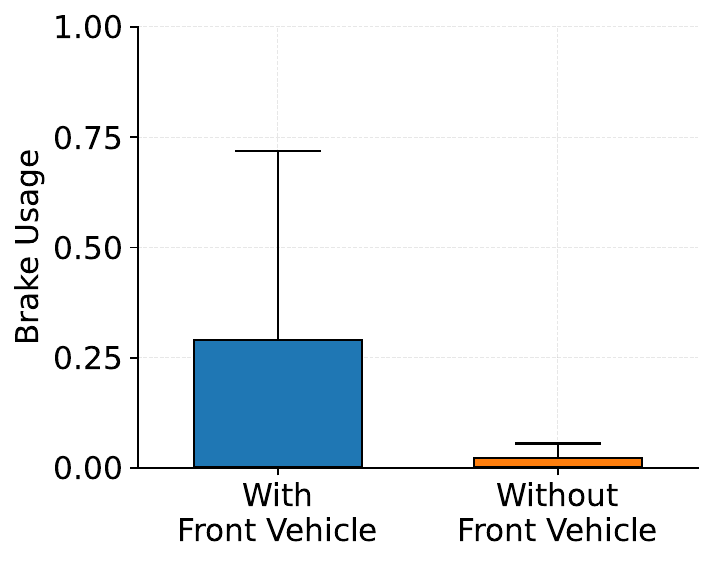} 
       \caption{Brake Usage}
   \end{subfigure}
   \hfill
   \begin{subfigure}[b]{0.235\textwidth}
       \centering\includegraphics[width=\textwidth]{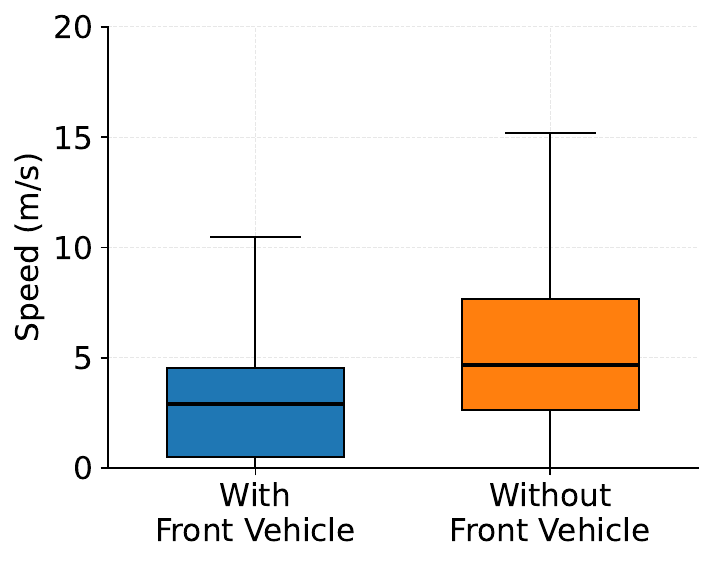}
       \caption{Speed}
   \end{subfigure}
   \hfill
   \begin{subfigure}[b]{0.235\textwidth}
       \centering\includegraphics[width=\textwidth]{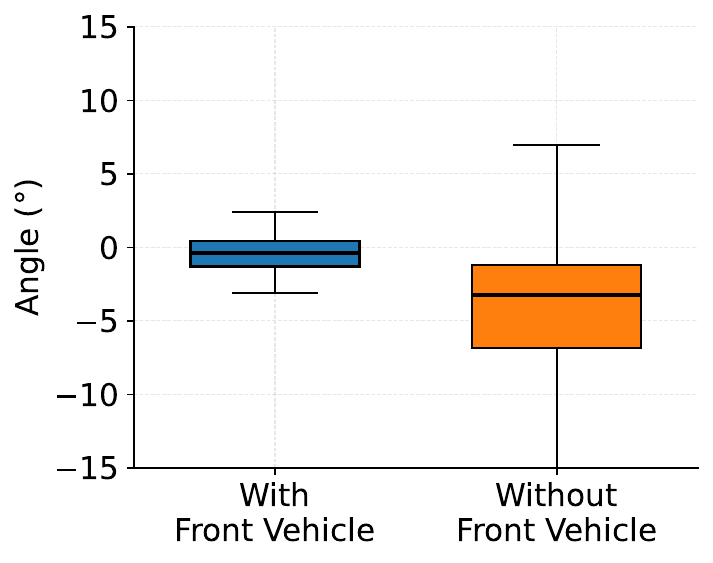} 
       \caption{Head Position}
   \end{subfigure}
    \caption{Comparison of drivers ' responses with and without a leading vehicle.}
        \label{fig:tb-context}
\end{figure*}

\begin{figure}[t]
   \centering
   \begin{subfigure}[b]{0.235\textwidth}
       \centering\includegraphics[width=\textwidth]{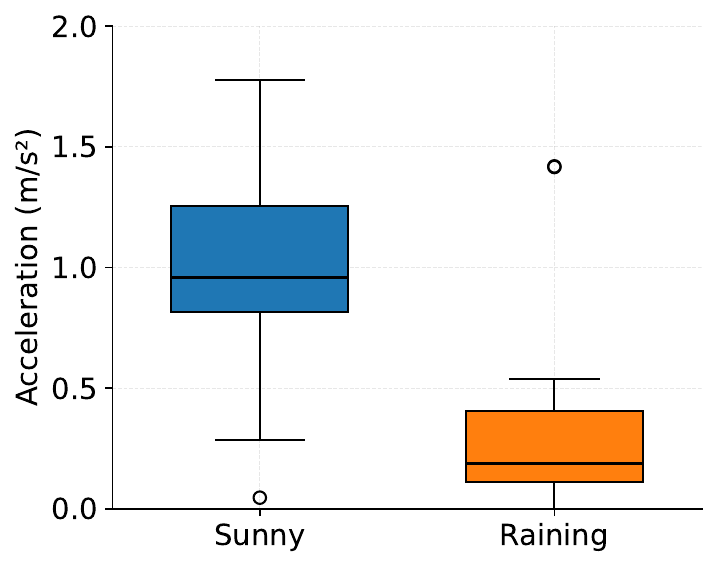}
       \caption{Accerlation}
   \end{subfigure}
   \hfill
   \begin{subfigure}[b]{0.235\textwidth}
       \centering\includegraphics[width=\textwidth]{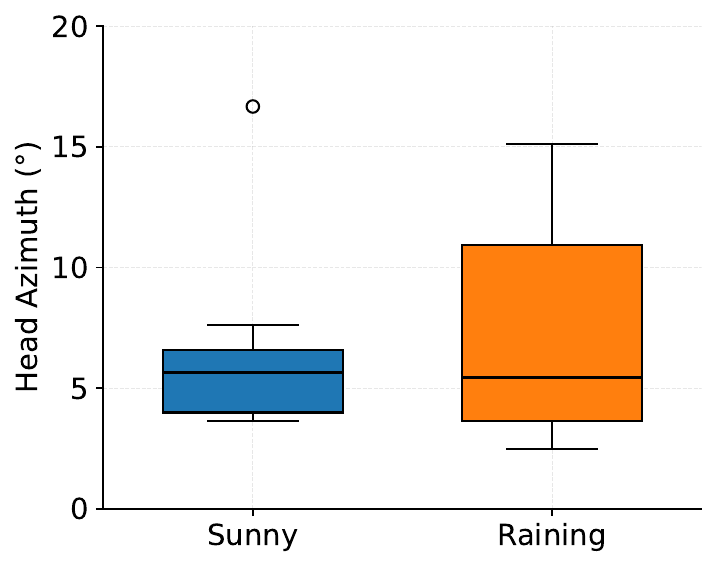} 
       \caption{Azimuth Angle Change}
   \end{subfigure}
    \caption{Driving under different weather conditions. }
    \label{fig:t-weather}
\end{figure}

\noindent \textbf{Remarks.}
Across both the real-world datasets, we show that the complexity of interpreting these signals is. 
Various personal driving styles and deliberate self-regulation strategies often mask or mimic cognitive difficulty, making it challenging to differentiate compensatory behavior from emerging impairment. 
Thus, extracting stable, safety-relevant meaning from them requires models that look beyond surface patterns to the underlying mechanisms driving behavioral change.


\subsection{Context and Environmental Awareness}
\label{sec:challenge2}
\noindent\textbf{How Does Context Shape Senior Driving Behavior?} 
Driving behavior cannot be interpreted in isolation. 
Its meaning depends heavily on the surrounding context. 
The same hard brake or sharp steering correction may reflect either a well-timed response to an external hazard or a delayed reaction to an anticipated event. 
For seniors, this distinction is crucial because age-related perceptual and attentional changes can alter the processing of environmental information. 
Our analysis seeks to quantify how their actions change as context shifts, revealing the extent to which external conditions amplify, suppress, or reshape age-related driving patterns.

To systematically examine this effect, we analyzed controlled simulator data in which both older and younger participants completed identical routes under two conditions: driving with and without a lead vehicle. 
This design isolates environmental context as the only varying factor. 
As shown in Figure~\ref{fig:tb-context}, the presence of a lead vehicle significantly alters senior driving behavior. 
When another car is directly ahead, older drivers consistently apply less throttle and more braking, maintaining substantially lower speeds compared to segments without a lead vehicle. 
This underscores a heightened emphasis on maintaining safety margins and avoiding uncertainty in closing speed.
Moreover, we analyzed their head direction in two scenarios.
Seniors exhibit reduced head-scanning range in these conditions.
Also, they are more concentrated around 0 degrees, indicating that they direct their visual attention primarily toward the vehicle in front rather than scanning the environment broadly.
Taken together, these results demonstrate that seniors' behavior is highly context-dependent and that understanding their safety requires analyzing how they adapt their control strategies to varying traffic conditions.

We further analyze how seniors adapt their driving behavior under different weather conditions. 
Our testbed includes both clear-sunny with high visibility and rainy with reduced contrast and impaired visual clarity. 
As shown in Figure~\ref{fig:t-weather}, seniors significantly reduce their acceleration when driving in rain, adopting a more conservative control strategy. 
The head-movement patterns also change.
The azimuth-angle change exhibits more frequent, wider scanning motions, indicating increased visual effort to compensate for reduced visibility. These observations highlight that weather is a potent modulator of senior driving behavior, reinforcing the need for context-aware sensing and modeling.

\noindent\textbf{Remark.} 
Our observations make clear that senior driving behavior cannot be understood solely from driver telemetry.
The surrounding environment fundamentally shapes how behaviors manifest. 
Seniors exhibit markedly different driving and body-movement patterns across environments, demonstrating that identical behaviors can signal differently in various contexts. 
Additionally, collecting detailed contextual information raises significant privacy concerns. 
Future deployments will therefore require privacy-preserving methods to prevent the exposure of sensitive location or behavioral data.
These findings highlight a critical takeaway: future assessment systems must move beyond driver-only monitoring and incorporate explicit environmental sensing to correctly determine whether a behavior reflects a compensatory strategy or an actual safety risk. 
A multimodal, context-aware, and privacy-preserving approach is therefore essential for building fair, reliable, and actionable intelligence for senior driving.

\begin{figure}[t]
  \centering
  \includegraphics[width=0.9\linewidth]{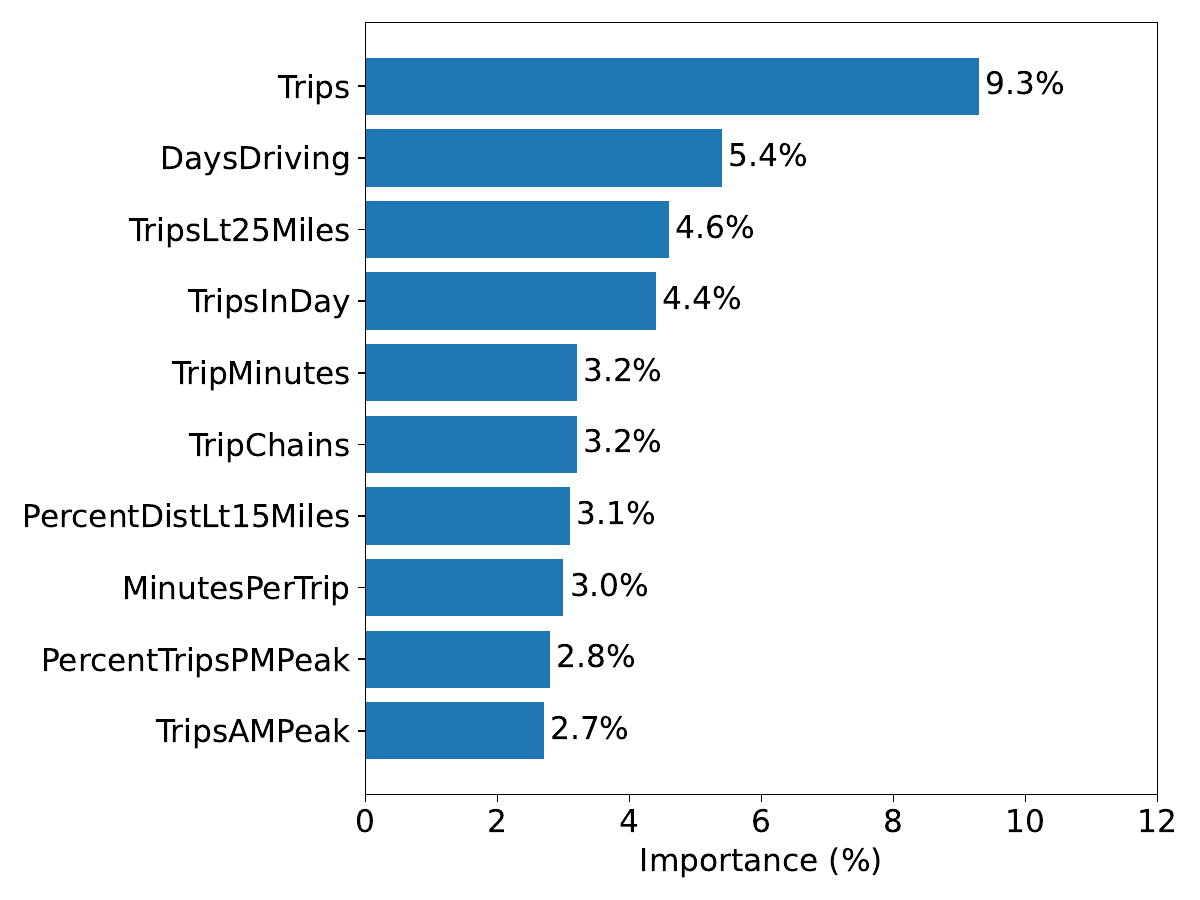}
  \caption{Safety Driving Importance Ratio}
  \label{fig:l_dp}
\end{figure}

\noindent \subsection{Explainable AI}
\label{sec:challenge3}
\textbf{How can we transform raw driving behavior into explanations that clinicians, policymakers, and older adults themselves can trust and act upon?}  
Explainability is not optional in this domain.
It is a prerequisite for any meaningful intervention.
Rather than depending solely on post-hoc interpretability techniques, cyber-physical control systems enforce safety through explicit physical models and bounded action spaces \cite{Proteus25Liu, Adonis25Liu, hydra24liu, wang2022detection,liu2025bench, gan2023poster}.
Decisions such as recommending cessation of driving, initiating cognitive screening, or prescribing adaptive training cannot rely on opaque statistical clusters or uninterpretable model scores. 
To achieve this, future work must extend our current descriptive framework toward causally informed interpretability by integrating multimodal sensing, constructing context-aware causal graphs, and validating hypotheses through controlled simulation. 
Only by converting behavioral fingerprints into mechanistic explanations can we build AI systems that are accurate, accountable, and truly actionable for real-world senior driving assessment.

We leverage the LongROAD dataset to develop a reproducible, interpretable unsupervised analysis pipeline designed to uncover latent behavioral phenotypes among older drivers and examine their association with driving ability. 
A subset of participants was manually annotated using clinical, self-reported, and family feedback to distinguish those who remain safe drivers from those who require further evaluation. 
Using these annotations as external references, we applied K-means clustering to all drivers to identify natural groupings that reflect distinct driving behavior profiles. The resulting framework produces interpretable outputs, including participants' segments and a ranked list of the top behavioral features, as shown in Figure \ref{fig:l_dp}.
It reveals that trip frequency, driving days, and short-distance trip ratios emerge as dominant indicators of age-related adaptation in mobility patterns.

While this workflow successfully extracts distinct behavioral signatures and quantifies their statistical contribution to group separation, it remains constrained by a fundamental limitation.
It discovers patterns but cannot explain the underlying mechanisms. 
The current model reveals how older adults drive differently, but it does not explain why they do so. 
For instance, a reduced number of trips could reflect lifestyle change rather than cognitive decline; fewer nighttime drives may signal deliberate risk aversion, impaired visual acuity, or contextual constraints such as limited transportation access. 
Without explicit modeling of driver intent, perceptual uncertainty, or environmental causality, the system risks conflating compensatory self-regulation with cognitive impairment, thereby limiting both its clinical interpretability and ethical deployability.

\noindent\textbf{Remark.} 
Our observations indicate that current approaches to driving analysis are insufficient for providing detailed, explainable insights into decision-making processes. A more comprehensive evaluation, incorporating causal inference, is needed to clarify the underlying factors influencing driving behavior. Such an approach can support interpretable explanations and inform future interventions, including senior-driver education programs, thereby enhancing road safety.



%% file: content/3_system_design.tex
\section{The \ours Framework}

\begin{table*}[ht]
\centering
\caption{\textbf{Concise summary of grand challenges, supporting patterns, system components, and solution directions.}}
\label{tab:grand_challenges_final}

\renewcommand{\arraystretch}{1.25}
\setlength{\aboverulesep}{0.6ex}
\setlength{\belowrulesep}{0.6ex}

\newcolumntype{C}[1]{>{\centering\arraybackslash}m{#1}}

\begin{tabular}{C{3.0cm} C{5.0cm} C{3.2cm} C{5.1cm}}
\specialrule{2pt}{0pt}{0pt}
\textbf{Grand Challenge} &
\textbf{Observations} &
\textbf{Vision Component} &
\textbf{Design Directions} \\
\specialrule{2pt}{0pt}{5pt}

\textbf{\makecell{From Raw Data \\to Senior Safety}} &
\makecell[l]{
• High intra-driver variability \\
• Inconsistent Behavior \\
• Compensatory behaviors 
} &
\textbf{\makecell{Multi-Scale\\ Representation\\ Learning}} &
\makecell[l]{
• Richer sensing  \\
• Longitudinal modeling \\
• Causal inference
}
\\
\dmidrule

\textbf{\makecell{Context \&\\ Environment\\ Awareness}} &
\makecell[l]{
• Scenario-dependent behavior \\
• Lead-vehicle influences \\
• Context avoidance
} &
\textbf{\makecell{Context-Aware\\ Perception \&\\ Fusion}} &
\makecell[l]{
• Multimodal context fusion \\
• Situation-normalized evaluation \\
• Real-time edge fusion
}
\\
\dmidrule

\textbf{\makecell{Explainable \&\\ Actionable AI}} &
\makecell[l]{
• Clusters show diff., not causes \\
• Clinical interpretability needs \\
• Sensitive Data
} &
\textbf{\makecell{Transparent\\ Actionable\\ Intelligent}} &
\makecell[l]{
• Concept-based interpretability\\
• Privacy Preservation  \\
• In-Vehicle Deployment
}
\\

\specialrule{2pt}{0pt}{5pt}
\end{tabular}
\end{table*}

With multi-scale observations, several key challenges arise in the analysis of senior driving. 
The first challenge is to move from raw data to senior safety. 
While seniors exhibit distinct driving patterns, accurately extracting these signatures is complicated by inherent behavioral variability and compensatory self-regulation. 
The second challenge is to transition from behavior to context awareness.
The environment intrinsically shapes driving analysis. 
The system must support robust multimodal integration with context to enable comprehensive analysis. 
Finally, the third challenge lies in the correlation between action and the other variables. 
Interpretability is not merely optional but essential, as these assessments directly influence the future independence and well-being of older adults.

In the face of these challenges, we envision \ours, a continuous, contextualized, and explainable AIoT framework designed to support safe driving for older adults through naturalistic driving analysis. 
The system advances this vision through three core components: multi-scale representation learning that captures both driving behavior and body movement, context-integrated modeling that interprets behavior in relation to situational demands, and transparent, actionable, valuable intelligence to drivers, caregivers, and clinicians. 

\subsection{Multi-Scale Representation Learning}
Modern vehicles generate continuous streams of throttle pressure, brake force, steering dynamics, GPS traces, and body movement. 
Interpreting these signals is inherently challenging while driving.
Senior driving introduces substantial natural variability in daily cognitive condition, making consecutive trips by the same driver fundamentally different and obscuring reliable indicators of functional change. 
Another complication arises from self-regulation.
Older adults adjust their driving by slowing down, avoiding complex routes, or limiting driving to familiar times of day. 
These compensatory strategies create a false appearance of safety and can obscure early signs of functional decline. 
Finally, driving is an overlearned skill shaped by decades of experience, producing highly individualized behavioral signatures that differ markedly across older adults. 
This individualized baseline makes it difficult to distinguish a meaningful cognitive change from a benign personal preference. Together, these factors undermine assumptions of behavioral consistency and highlight the need for longitudinal, personalized modeling rather than population-level or single-trip analyses.

To capture this whole spectrum, \ours learns multi-scale representations that fuse fast control actions with fine-grained motion features and long-term behavioral regularities.
At the lowest time scale, the system encodes rapid dynamic behavior, including steering movements, pedal pressure, head turns, gaze shifts, and micro-postural movements, by applying temporal convolution and self-attention modules to raw sensor streams. 
These encoders transform heterogeneous signals into aligned features that describe moment-to-moment control responses.
At the intermediate scale, \ours aggregates these features into short behavioral segments that capture how driving unfolds within situational constraints. 
Segment-level encoders then learn patterns of occurrence between body movement and vehicle actions, enabling the system to detect when a driver compensates for cognitive load or reacts to road conditions.
At longer horizons, \ours constructs daily and weekly profiles that summarize routine driving habits, typical head–body coordination patterns, and stable control strategies. 
A hierarchical temporal model links these levels, enabling rapid adjustments to be interpreted within the context of broader mobility rhythms. 
This structure enables the system to distinguish short, benign fluctuations from slow and consistent behavioral drift.
By combining learning from fine-grained features and long-range behavioral structure, \ours produces stable, interpretable representations that support continuous monitoring and reliable long-term assessment.

\subsection{Context-Integrated Modeling}
Driving behavior is powerfully shaped by the surrounding environment, making context fusion essential for accurate inference.  
Without a contextual understanding, models risk misclassifying safe, adaptive behaviors as impairments.  
A comprehensive system must integrate multimodal data sources, including traffic density, weather conditions, road geometry, and visual perception, to reconstruct the surrounding driving environment and assess behavioral appropriateness. 
Multimodal fusion remains an open problem.
Data are asynchronous, incomplete, and often lack ground truth for causal attribution. 
Progress demands new architectures that reason jointly over driver intent and environmental state, supported by shared datasets and benchmark protocols for real-world validation.

\ours embeds environmental information directly into its behavioral reasoning pipeline through a structured, multi-stage fusion architecture shown in Table \ref{tab:grand_challenges_final}. 
Environmental inputs are first encoded through modality-specific encoders that extract spatial and temporal features. 
In parallel, driving behavior signals are transformed into a unified action embedding.
To integrate these heterogeneous streams, \ours employs cross-modal alignment layers that synchronize environmental features with driver actions at shared timestamps. Attention-based fusion modules then learn the interaction patterns between drivers' perception and their responses. 
This combined representation enables \ us to reason about driving as contextually grounded behavior rather than isolated kinematics. 
As a result, the fused model produces route-independent, environmentally fair, and more tightly linked evaluations to the underlying functional capability. 
This context-aware reasoning forms a reliable basis for detecting actual decline rather than environment-induced variation.

\subsection{Transparent and Actionable Intelligence}
Driving assessments directly influence a person's legal status, daily independence, and long-term mobility. 
Opaque models are therefore unacceptable. 
Clinicians, regulators, and older drivers must be able to trust the system's conclusions, which requires transparent reasoning that clearly links observed driving behavior to the cognitive processes that support safe operation.
To ensure \ours is not limited to theoretical utility, we must bridge the gap to practice. The transition from algorithmic design to real-world deployment, however, imposes equally rigorous operational constraints.
In-vehicle hardware provides limited compute and memory resources, and inference must occur with minimal latency to ensure safety during dynamic driving events. 
At the same time, privacy requirements are stringent.
Sensitive information about daily routines and health-related patterns must remain on the vehicle and cannot be transmitted externally. 
These constraints call for models that are compact enough for edge execution, robust to noisy, naturalistic conditions, and capable of generating interpretable, defensible outputs rather than opaque predictions. 
The overarching challenge is to design an assessment pipeline that is actionable for users while remaining strictly private, resource-efficient, and dependable in everyday driving environments.
 
\ours embodies this principle through an integrated functional reasoning layer that translates high-dimensional, noisy driving traces into transparent, cognitively grounded explanations. Unlike traditional end-to-end models that output opaque risk scores, our system adopts a hierarchical ``Semantic Abstraction'' approach co-designed with an expert panel of geriatricians and neuropsychologists.
This process follows a three-stage interpretability pipeline.
We first apply semantic abstraction.
The system first decomposes raw sensor data into granular micro-behaviors.
This step removes sensor noise and isolates behavioral primitives. The second step is Causal Mapping to Cognitive Domains.
These micro-behaviors are then mapped to specific cognitive domains based on clinical rubrics established by our medical partners. 
For instance, a recurrent failure to maintain lane position is not flagged merely as ``bad driving'', but explicitly correlated with potential lapses in spatial attention. 
The final step is diagnostic reporting. 
Instead of a binary classification, \ours generates a structured Preliminary Diagnostic Report. 
This report mimics the logic of a clinical assessment, providing readable evidence chains.
This ensures the output is auditable by physicians and defensible in a caregiving context.

Moving from the paper to the real world necessitates a rigorous approach to deployment. 
The central challenge lies in executing these complex models within resource-constrained environments. To address this, our deployment strategy mitigates critical latency and privacy bottlenecks through targeted architectural optimizations.
We implement a heterogeneous computing pipeline where sensor streams are aligned and filtered using lightweight Digital Signal Processing modules before reaching the neural networks. 
This ensures that only high-utility data consumes GPU cycles.
To fit large reasoning models onto automotive-grade chips, we employ a workflow that distills and then quantizes. 
The reasoning layer is distilled from large server-side teachers into compact student models, which are then pruned and quantized to lower precision. 
This enables fast inference, which is critical for real-time safety interventions. 
Recognizing the sensitivity of medical and location data, \ours keeps all raw video and GPS history locally on the vehicle. 
The system evolves via Federated Learning (FL), transmitting only sparse, differentially private gradient updates to the cloud. This allows the global model to learn from population-wide trends while mathematically guaranteeing that individual user trajectories can never be reconstructed.

\subsection{Broader Impact}
By systematically addressing these challenges, \ours outlines a viable roadmap toward next-generation support for older drivers. 
Its architectural principles enable continuous monitoring that preserves privacy, personalized assessments that adapt to individuals' evolving abilities, and interpretable insights that guide meaningful interventions. 
Translating these technical capabilities into societal impact, \ours empowers clinicians with objective diagnostics and guides families through complex driving choices. 
This establishes a virtuous cycle of balancing individual autonomy with public safety through continuous, intelligent oversight.
Finally, our vision extends beyond technology to community service.
We aim to use human-centered AIoT to build a more inclusive, warmer society where every senior can age with dignity and independence, supported by a system that is as compassionate as it is intelligent.

%% file: content/6_related_work.tex
\section{Go Beyond \ours}
A senior driving assessment offers a unique opportunity to observe how driving ability evolves with age. 
Beyond \ours, we outline three forward-looking research directions that define the technological roadmap for next-generation aging and driving safety systems, each addressing gaps that current solutions cannot fill.

\subsection{Elderly Cognitive World Models} 
Rapid population aging has made cognitive impairment one of the most pressing challenges for sustaining independent living in later life.
This necessitates tools to characterize how cognition changes over months and years. 
Understanding these long-term cognitive trajectories is essential for designing support systems that preserve safety, autonomy, and quality of life.
This motivates the development of cognitive world models that explicitly represent how older adults interpret situations, form expectations, and choose actions as their abilities shift. 
Naturalistic driving offers a compelling channel for learning such models.
Driving places sustained demands on attention, planning, spatial reasoning, and hazard anticipation, all of which are sensitive to age-related cognitive changes. 
These demands make driving a setting in which early signs of cognitive difficulty manifest more clearly and consistently than in many daily activities, allowing differences in perception, decision-making, and reaction patterns to be captured directly through driving behavior.
A driving-based cognitive world model captures the internal decision structure that generates observable driving actions.
It enables systems to detect emerging difficulties, distinguish intrinsic decline from situational effects, and deliver personalized support without intruding on independence. 
This makes world-model learning a promising and technically grounded direction for advancing safe and sustainable mobility in an aging society.

\subsection{Digital Therapeutics and Active Health}
As vehicles transition from passive transport platforms to embodied intelligent systems, they offer a unique opportunity to deliver continuous, real-world cognitive support for aging drivers. 
Driving is a high-demand task, making it a natural substrate for in-situ therapeutic stimulation. 
A future in-vehicle system can move beyond monitoring to deliver just-in-time adaptive interventions grounded in neuroplasticity, using short, context-aware prompts or subtle adjustments to navigation cues to exercise core cognitive functions during low-risk segments of a trip. 
This establishes a form of in-vehicle cognitive digital therapeutics that adapts to each driver's trajectory. 
Sustained effectiveness, however, depends on trust and adherence. 
Vehicles must therefore evolve into socially competent agents that perceive drivers' communication in supportive, non-judgmental language and provide transparent explanations for their feedback. By combining moment-to-moment therapeutic opportunities with empathetic, human-centered interaction, this direction positions the vehicle as an active cognitive clinic that promotes healthier aging through everyday mobility.

\subsection{Federated Embodied Learning}
Modern vehicles now function as embodied intelligent agents, continuously closing the loop between perception, action, and feedback as they move through the physical world.
Federated learning in this setting must therefore operate on embodied interaction rather than static datasets.
Each update reflects how a vehicle senses its environment, executes control, and adapts to a driver's behavior. 
This creates a distinct FL regime in which heterogeneity arises from differences in road conditions, sensing quality, vehicle dynamics, and driver characteristics, rather than from simple variation in data distribution. 
Future systems must design FL algorithms that preserve the structure of embodied experience by aligning representations that couple sensing and actuation. 
Communication should focus on transmitting compact summaries of evolving interaction dynamics to avoid unnecessary bandwidth usage and to protect user privacy. 
Strong differential privacy is essential to prevent updates from revealing sensitive behavioral or health information embedded in control traces. 
By unifying federated learning with embodied intelligence, vehicles can collaboratively improve a shared foundation model grounded in real-world interaction, and each car retains its own raw experience.

%% file: content/7_conclusion.tex
\section{Conclusion}
This paper presents a systems vision for using naturalistic driving as a continuous window into how driving ability changes with age. 
Every day driving captures rich signals about how older adults respond to traffic, navigate complex roads, and manage routine behavior.
\ours offers a unified framework for turning these daily driving patterns into a long-term view of functional driving ability by combining richer in-vehicle sensing, multi-scale behavioral modeling, and context-aware analysis within a privacy-preserving, edge-first design. Although \ours is an early step, it highlights a growing opportunity to use vehicles themselves as long-term monitors that support safer driving in later life. 
Realizing this vision will require progress in multimodal sensing, practical edge deployment, fair and interpretable assessment methods, and responsible data practices. 
By outlining these challenges and opportunities, this paper seeks to guide a research agenda that supports safe independence for an aging population. 